\documentclass[12pt]{article}

\usepackage[margin=1.37in]{geometry}
\usepackage[utf8]{inputenc}
\usepackage[T1]{fontenc}
\usepackage{url} 
\usepackage{booktabs} 
\usepackage{amsfonts} 
\usepackage{nicefrac}  
\usepackage{microtype}  
\usepackage{graphicx}

\usepackage{amsmath}
\usepackage{amssymb}
\usepackage{multirow}
\usepackage{bbm}
\usepackage{enumitem}
\usepackage{wrapfig}
\usepackage{epsfig,psfrag}
\usepackage{caption}
\usepackage{subcaption}
\usepackage[normalem]{ulem}
\usepackage{bm}
\usepackage{authblk}
\usepackage{natbib}
\usepackage{changepage}
\usepackage{bbold}

\usepackage{amsthm}
\theoremstyle{plain}

\usepackage{hyperref}  
\hypersetup{colorlinks,
citecolor=blue,
filecolor=blue,
linkcolor=red,
urlcolor=blue,
pdftex}
\usepackage[capitalize,noabbrev]{cleveref}

\title{
\begin{adjustwidth}{-.7in}{-.7in}
\centering \large \textbf{Topological Classification in a Wasserstein Distance Based Vector Space}
\end{adjustwidth}
}

\author[1+]{\normalsize Tananun Songdechakraiwut}
\author[2+]{Bryan M. Krause}
\author[2,3+]{Matthew I. Banks}
\author[4]{\\Kirill V. Nourski}
\author[1+]{Barry D. Van Veen}
\date{}
\affil[1]{\small Department of Electrical and Computer Engineering}
\affil[2]{\small Department of Anesthesiology}
\affil[3]{\small Department of Neuroscience}
\affil[+]{\small University of Wisconsin--Madison}
\affil[4]{Department of Neurosurgery, Iowa Neuroscience Institute, University of Iowa}

\begin{document}

\maketitle

\begin{abstract}
Classification of large and dense networks based on topology is very difficult due to the computational challenges of extracting meaningful topological features from real-world networks. In this paper we present a computationally tractable approach to topological classification of networks by using principled theory from persistent homology and optimal transport to define a novel vector representation for topological features. The proposed vector space is based on the Wasserstein distance between persistence barcodes. The 1-skeleton of the network graph is employed to obtain 1-dimensional persistence barcodes that represent connected components and cycles. These barcodes and the corresponding Wasserstein distance can be computed very efficiently. The effectiveness of the proposed vector space is demonstrated using support vector machines to classify simulated networks and measured functional brain networks.
\end{abstract}

\newpage

\section{Introduction}
Networks are ubiquitous representations for describing complex, highly interconnected systems that capture potentially intricate patterns of relationships between nodes. 
 \citep{barrat2004architecture}.
Finding meaningful characterizations of network structure is very difficult, especially for large and dense networks with node degrees ranging over multiple orders of magnitude \citep{bullmore2009complex,honey2007network}.

Persistent homology \citep{barannikov1994framed,edelsbrunner2000topological,wasserman2018topological} is an emerging tool for understanding, characterizing and quantifying the topology of complex networks \citep{carriere2020perslay,songdechakraiwut2021topological}.
Topology is characterized using connected components (0-dimensional topological features), cycles (1-dimensional topological features), voids (2-dimensional topological features) and higher dimensional, difficult to visualize, objects.
Connected components and cycles are the most dominant and fundamental topological features of real networks.
Many networks naturally organize into modules or connected components \citep{bullmore2009complex,honey2007network}. Similarly, cycle structure is ubiquitous and is often interpreted in terms of information propagation, redundancy and feedback loops \citep{keizer1995insp3,kwon2007analysis,ozbudak2005system,venkatesh2004multiple,weiner2002ptdinsp}. 
While voids have recently been considered in cosmology, they tend to be relatively rare, thus reducing their discriminative power \citep{biagetti2021persistence}.
Anything of higher dimension than the void is beyond the apparent physical space of how we see the world, and thus is difficult to understand and interpret. This motivates use of only connected components and cycles to differentiate networks.

Topological features are represented using descriptors called \emph{persistence barcodes} \citep{ghrist2008barcodes}.
Effective use of such topological descriptors requires a notion of proximity, that is, a metric that quantifies the distance between persistence barcodes.
However, incorporating barcodes into learning tasks requires more than identifying a suitable distance metric.
In general persistence barcodes do not have the algebraic structures required by a large class of learning methods such as support vector machines (SVMs).
For example, it is unclear whether elementary operations such as addition, scalar multiplication, and inner product have analogues in the space of persistence barcodes.

Approaches that embed persistence barcodes into a space with adequate structure such as vector spaces or Hilbert spaces have recently been proposed to address this limitation. For example,
\citet{adams2017persistence,bubenik2015statistical,carriere2015stable} describe methods for extracting vector representations from persistence barcodes. Other authors \citep{carriere2017sliced,kusano2016persistence,reininghaus2015stable} define implicit feature representations in a Hilbert space using the kernel trick \citep{hofmann2008kernel}.
However, none of these methods preserves the underlying distance in the original space of persistence barcodes \citep{carriere2019metric}. 
Recently, it was shown that persistence barcodes are inherently 1-dimensional if the network topology is limited to connected components and cycles \citep{songdechakraiwut2021topological}.

Motivated by this result, here we present a novel \emph{topological vector space} (TopVS) for 1-dimensional persistence barcodes.
The $p$-norm distance in TopVS is equivalent to the $p$-Wasserstein distance in the original space of persistence barcodes.
This equivalence allows the computation of summary statistics such as the mean of persistence barcodes to be easily performed in TopVS.
The utility of TopVS is demonstrated in a SVM-based classification task.
Statistical validation is used to demonstrate the effectiveness of TopVS relative to competing approaches when discriminating subtle topological features in simulated networks.
TopVS is further illustrated by classifying measured functional brain networks associated with different levels of arousal during administration of general anesthesia. TopVS performs very well compared to other topology-based approaches in both the simulated and measured data.

The paper is organized as follows. Background on the 1-dimensional representation of persistence barcodes is given in \cref{sec:prelim}, while \cref{sec:method} presents our Wasserstein distance based TopVS for 1-dimensional persistence barcodes. 
In \cref{sec:sim,sec:brain}, simulated and measured networks are used to compare the classification performance of TopVS relative to that of several baseline methods. \cref{sec:conclusion} concludes the paper with a brief discussion of the potential impact of this work.

\section{One Dimensional Persistence Barcodes}
\label{sec:prelim}

\subsection{Graph Filtration}

\begin{figure}[t]
\centering
\centerline{\includegraphics[width=1.15\linewidth]{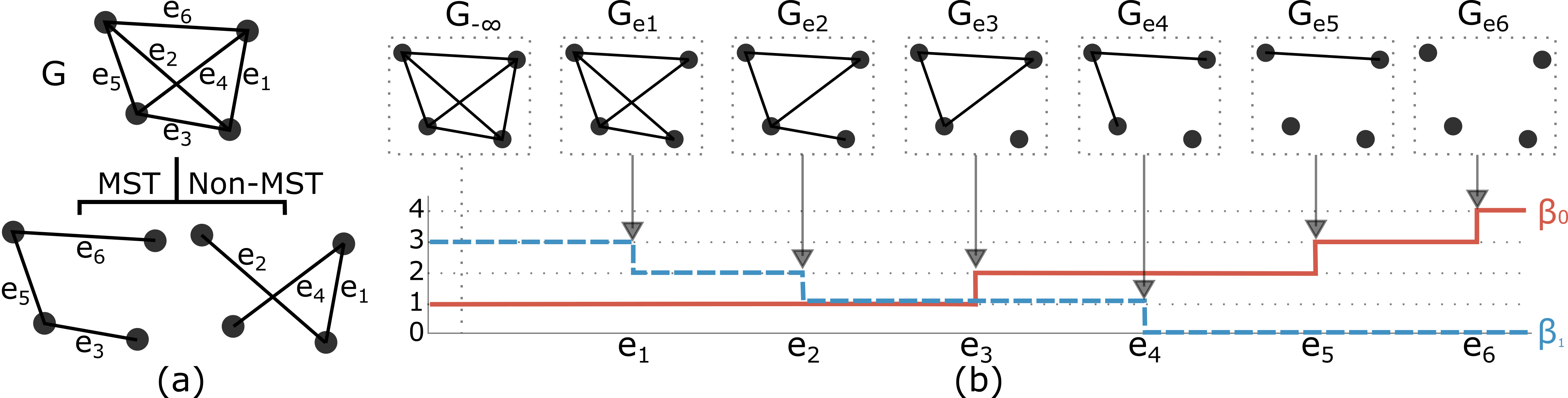}}
\caption{(a) Four-node network $G$ decomposes into its maximum spanning tree (MST) and a subnetwork with non-MST edge weights. (b) As the filtration value increases, the number of connected components $\beta_0$ monotonically increases while the number of cycles $\beta_1$ monotonically decreases. Connected components are born at the MST edge weights $e_3,e_5,e_6$ while cycles die at the non-MST edge weights $e_1,e_2,e_4$.}
\label{fig:schematic}
\end{figure}

Define a network as an undirected weighted graph $G=(V,\bm{w})$ with a set of nodes $V$, and a weighted adjacency matrix $\bm{w}=(w_{ij})$. The number of nodes is denoted by $|V|$.
Define a binary graph $G_\epsilon$ with the identical node set $V$ by thresholding the edge weights so that an edge  between nodes $i$ and $j$ exists if $w_{ij} > \epsilon$. The binary graph is viewed as a simplicial complex consisting of only nodes and edges, that is, a 1-skeleton \citep{munkres2018elements}.
As $\epsilon$ increases, more and more edges are removed from the network $G$. Thus, we have a nested sequence of 1-skeletons:
\begin{equation}
\label{eq:graphfiltration}
G_{\epsilon_0} \supseteq G_{\epsilon_1} \supseteq \cdots \supseteq G_{\epsilon_k} ,
\end{equation}
where $\epsilon_0 \leq \epsilon_1 \leq \cdots \leq \epsilon_k$ are called filtration values. This sequence of 1-skeletons is called a \emph{graph filtration} \citep{lee2012persistent}.
\cref{fig:schematic} illustrates the graph filtration of a four-node network.
Note that other filtrations for analyzing graphs have been proposed based on descriptor functions such as heat kernels \citep{carriere2020perslay} and task-specific learning \citep{hofer2020graph}, in contrast to the use of edge weights.

\subsection{Birth-death Decomposition}
\label{sec:decomposition}

Persistent homology keeps track of the birth and death of topological features over filtration values $\epsilon$. A topological feature that is born at a filtration $b_i$ and persists up to a filtration $d_i$, is represented as a 2-dimensional point $(b_i,d_i)$ in a plane. A set of all the points $\{(b_i,d_i)\}$ is called \emph{persistence diagram} \citep{edelsbrunner2008persistent} or, equivalently, \emph{persistence barcode} \citep{ghrist2008barcodes}. In the 1-skeleton, the only non-trivial topological features are connected components (0-dimensional topological features) and cycles (1-dimensional topological features). There are no higher-dimensional topological features in the 1-skeleton, in contrast to clique complexes \citep{otter2017roadmap,zomorodian2010fast} and Rips complexes \citep{ghrist2008barcodes}. This simplifies the persistence barcode and significantly reduces the computational complexity of the corresponding topological analysis.

The graph filtration given in (\ref{eq:graphfiltration}) begins with a complete graph $G_{-\infty}$, sequentially removes edges at higher filtration values $\epsilon$, and arrives at an edgeless graph $G_{+\infty}$.
As $\epsilon$ increases, the number of connected components $\beta_0(G_{\epsilon})$ and cycles $\beta_1(G_{\epsilon})$ are monotonically increasing and decreasing, respectively \citep{songdechakraiwut2021topological}. Specifically, $\beta_0(G_\epsilon)$ increases from the complete graph consisting of a single connected component $\beta_0(G_{-\infty}) = 1$ to the node set $\beta_0(G_{\infty}) = |V|$. There are $\beta_0(G_{\infty}) - \beta_0(G_{-\infty}) = |V| - 1$ connected components that are born over the filtration. 

Once connected components are born, they will remain until the node set $G_{+\infty}$ is reached, so all death values are at $+\infty$. Thus, the representation of the connected components can be simplified to a collection of sorted birth values $B(G) = \{b_i \}_{i=1}^{|V|-1}$. On the other hand, all cycles are born with the complete graph $G_{-\infty}$ and thus have birth values at $-\infty$. Again we can simplify the representation of the cycles as a collection of sorted death values $D(G)=\{ d_i \}$.
The removal of an edge must result in either the birth of a connected component or the death of a cycle. Thus every edge weight must also be in either $B(G)$ or $D(G)$, resulting in the decomposition of the edge weight set $W=\{w_{ij}\,|\,i>j\}$ into $B(G)$ and $D(G)$ \citep{songdechakraiwut2021topological}.
Since the complete graph $G_{-\infty}$ has $\frac{|V|(|V|-1)}{2}$ edge weights, the number of cycles in $G_{-\infty}$ is equal to $\frac{|V|(|V|-1)}{2} - (|V|-1) = 1 + \frac{|V| (|V| - 3)}{2}$.
Thus, every different network with the same node size $|V|$ has a birth set $B$ and a death set $D$ of the same cardinality as $|V| - 1$ and $1 + \frac{|V| (|V| - 3)}{2}$, respectively.
Note that other filtrations \citep{carriere2020perslay,ghrist2008barcodes,hofer2020graph,otter2017roadmap,petri2013topological,zomorodian2010fast} do not necessarily share this monotonicity property. Thus, their persistence barcodes are not 1-dimensional, and the number of points in the persistence barcodes may vary for different networks of the same size.

$B(G)$ comprises edge weights in the \emph{maximum spanning tree} (MST) of $G$ \citep{lee2012persistent}, and can be computed using standard methods such as Kruskal's  \citep{kruskal1956shortest} and Prim's algorithms \citep{prim1957shortest}.
Once $B(G)$ is identified, $D(G)$ is given as the remaining edge weights that are not in the MST. Thus $B(G)$ and $D(G)$ are computed very efficiently in $O(|V|^2\log |V|)$ operations.
The example network of \cref{fig:schematic} has $B(G)=\{e_3,e_5,e_6\}$ and $D(G)=\{e_1,e_2,e_4\}$.

\section{Topological Space with Wasserstein Distance}
\label{sec:method}

\subsection{Wasserstein Distance Simplification}

The graph filtration given in (\ref{eq:graphfiltration}) results in significant simplification of the Wasserstein distance between barcode descriptors of networks.
Let $\bm X$ and $\bm Y$ be 2-dimensional random vectors describing the stochastic nature of points in conventional 2-dimensional persistence barcodes \citep{ghrist2008barcodes}. Let $\phi_{\bm{X}}$ and $\phi_{\bm{Y}}$ be probability distributions of $\bm X$ and $\bm Y$, respectively. The $p$-Wasserstein distance between $\phi_{\bm{X}}$ and $\phi_{\bm{Y}}$ is defined as \citep{kolouri2017optimal}
\begin{align}
\label{eq:2dwass}
    W_p(\phi_{\bm{X}},\phi_{\bm{Y}}) := \Big( \inf_{\phi_{\bm{X},\bm{Y}} \in \Phi} \int_{\mathbb{R}^2} (||x - y||_p)^p \,d\phi_{\bm{X},\bm{Y}}(x,y)\Big)^{1/p},
\end{align}
where $||\cdot||_p$ denotes the $p$-norm, and the infimum is taken over all joint distributions $\Phi$ of the random vectors $\bm X$ and $\bm Y$ with marginal distributions $\phi_{\bm{X}}$ and $\phi_{\bm{Y}}$. 
Intuitively, each distribution is a unit mass, and the Wasserstein metric measures the optimal transport plan with the minimum work of turning one mass into the other by moving points over $p$-norm induced distance.

Solving the optimization problem in (\ref{eq:2dwass}) is computationally costly and multiple approximation algorithms have been proposed to manage its computational complexity (see, e.g., \citep{cuturi2013sinkhorn,kerber2017geometry,Lacombe2018LargeSC,rabin2011wasserstein}). Consequently, use of Wasserstein distance with the 2-dimensional persistence barcodes resulting from conventional filtrations \citep{carriere2020perslay,ghrist2008barcodes,hofer2020graph,otter2017roadmap,petri2013topological,zomorodian2010fast} is computationally challenging.

On the other hand, the Wasserstein distance between the 1-dimensional barcodes of the graph filtration defined in (\ref{eq:graphfiltration}) can be obtained using a closed-form solution.
Let $G_1$ and $G_2$ be two given networks possibly with different node sizes, i.e., their birth and death sets may vary in size. Their underlying probability density functions on the persistence barcodes for connected components are defined in the form of Dirac masses \citep{turner2014frechet}:
\begin{align}
    f_{G_1,B}(x) &:= \frac{1}{|B(G_1)|} \sum_{b \in B(G_1)} \delta(x-b), \\
    f_{G_2,B}(x) &:= \frac{1}{|B(G_2)|} \sum_{b \in B(G_2)} \delta(x-b),
\end{align}
where $\delta(x-b)$ is a Dirac delta centered at the point $b$. Then the empirical distributions are the integration of $f_{G_1,B}$ and $f_{G_2,B}$ as
\begin{align}
    F_{G_1,B}(x) &= \frac{1}{|B(G_1)|} \sum_{b \in B(G_1)} \mathbb{1}_{b \leq x}, \\
    F_{G_2,B}(x) &= \frac{1}{|B(G_2)|} \sum_{b \in B(G_2)} \mathbb{1}_{b \leq x},
\end{align}
where $\mathbb{1}_{b \leq x}$ is an indicator function taking the value 1 if $b \leq x$, and 0 otherwise.
A pseudoinverse of $F_{G_1,B}$ is defined as
\begin{equation}
    F_{G_1,B}^{-1}(z) = \inf \{b \in \mathbb{R}\,|\, F_{G_1,B}(b) \geq z\},
\end{equation}
i.e., $F_{G_1,B}^{-1}(z)$ is the smallest $b$ for which $F_{G_1,B}(b) \geq z$.
Similarly, we define a pseudoinverse of $F_{G_2,B}$ as
\begin{equation}
    F_{G_2,B}^{-1}(z) = \inf \{b \in \mathbb{R}\,|\, F_{G_2,B}(b) \geq z\}.
\end{equation}
Then the empirical Wasserstein distance for connected components has a closed-form solution in terms of these pseudoinverses as \citep{kolouri2017optimal}
\begin{equation}
\label{eq:numerint}
    W_{p,B}(G_1,G_2) = \Big(\int_0^1 |F^{-1}_{G_1,B}(z) - F^{-1}_{G_2,B}(z)|^p\,dz\Big)^{1/p}.
\end{equation}
Similarly, the Wasserstein distance for cycles $W_{p,D}(G_1,G_2)$ is defined in terms of empirical distributions for death sets $D(G_1)$ and $D(G_2)$.

The empirical Wasserstein distances $W_{p,B}$ and $W_{p,D}$ are approximated by computing the Lebesgue integration in (\ref{eq:numerint}) numerically as follows.
Let 
\begin{align*}
    \widehat B(G_1) &= \{F^{-1}_{G_1,B}(1/m), F^{-1}_{G_1,B}(2/m), ..., F^{-1}_{G_1,B}(m/m)\}, \\
    \widehat D(G_1) &= \{ F^{-1}_{G_1,D}(1/n), F^{-1}_{G_1,D}(2/n), ..., F^{-1}_{G_1,D}(n/n)\}
\end{align*}
be pseudoinverses of network $G_1$ sampled with partitions of equal intervals.
Let $\widehat B(G_2)$ and $\widehat D(G_2)$ be sampled pseudoinverses of network $G_2$ with the same partitions of $m$ and $n$, respectively. Then the approximated Wasserstein distances are given by
\begin{align}
    \label{eq:1dwasscc}
    \widehat W_{p,B}(G_1,G_2) &= \Big( \frac{1}{m^p} \sum_{k=1}^m \big|F^{-1}_{G_1,B}(k/m) - F^{-1}_{G_2,B}(k/m)\big|^p \Big)^{1/p}, \\
    \label{eq:1dwasscycle}
    \widehat W_{p,D}(G_1,G_2) &= \Big( \frac{1}{n^p} \sum_{k=1}^n \big|F^{-1}_{G_1,D}(k/n) - F^{-1}_{G_2,D}(k/n)\big|^p \Big)^{1/p}.
\end{align}

\begin{figure}
\centering
\centerline{\includegraphics[width=.7\linewidth]{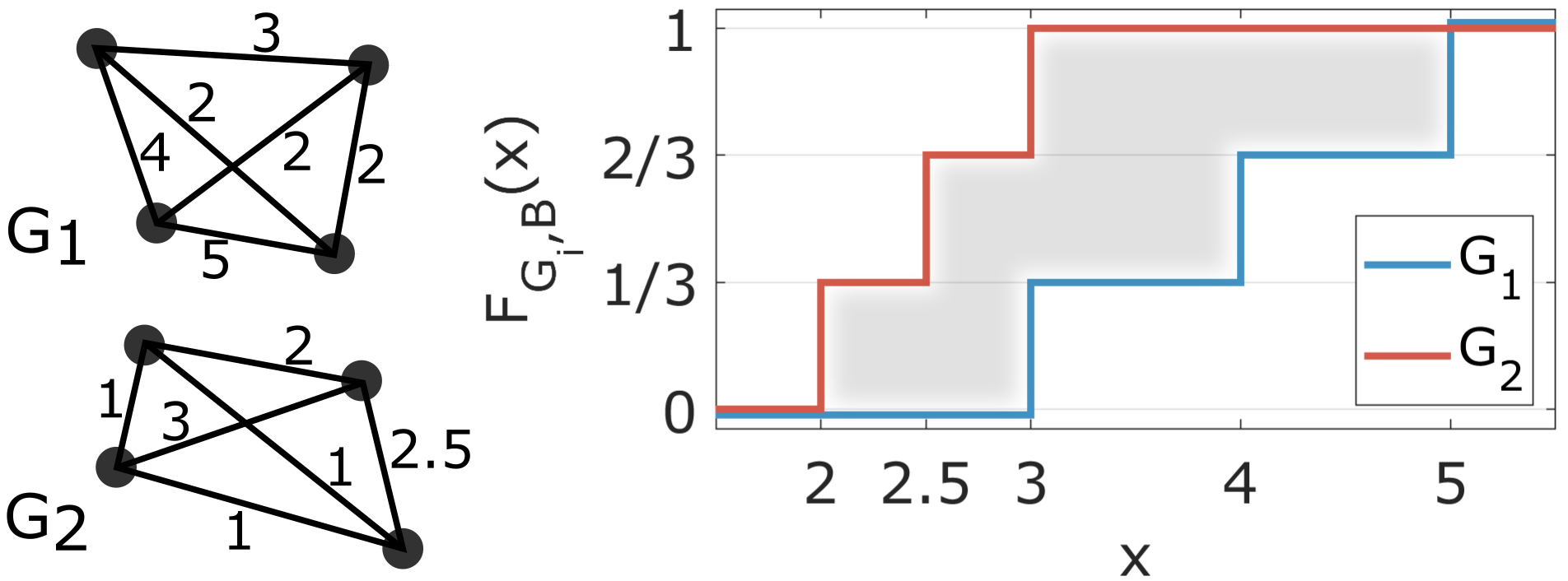}}
\caption{Four-node networks $G_1$ and $G_2$ have birth sets of connected components as $B(G_1)=\{3,4,5\}$ and $B(G_2)=\{2,2.5,3\}$, respectively. The 1-Wasserstein distance $W_{1,B}(G_1,G_2)= \frac{1}{3}\big( |3-2| + |4-2.5| + |5-3| \big),$ which is exactly equal to the shaded area situated between two empirical distributions $F_{G_1,B}$ and $F_{G_2,B}$.}
\label{fig:approxwass}
\end{figure}

For the special case when networks $G_1$ and $G_2$ have the same number of nodes, the exact computation of the Wasserstein distance is achieved using the original birth and death sets $B(G_1), B(G_2), D(G_1),$ and $D(G_2)$ as \citep{songdechakraiwut2021topological}
\begin{align}
    W_{p,B}(G_1,G_2) &= \Big( \frac{1}{|B(G_1)|^p} \sum_{b \in B(G_1)} |b - \tau_0^*(b)|^p \Big)^{1/p}, \\
    W_{p,D}(G_1,G_2) &= \Big( \frac{1}{|D(G_1)|^p} \sum_{d \in D(G_1)} |d - \tau_1^*(d)|^p \Big)^{1/p},
\end{align}
where $\tau_0^*$ maps the $l$-th smallest birth value in $B(G_1)$ to the $l$-th smallest birth value in $B(G_2)$, and $\tau_1^*$ maps the $l$-th smallest death value in $D(G_1)$ to the $l$-th smallest death value in $D(G_2)$, for all $l$.
The exact Wasserstein distances $W_{p,B}$ and $W_{p,D}$ are well-defined since the bijective mapping $\tau_0^*$ between same-cardinality sets of births is well-defined, as does $\tau_1^*$ for same-cardinality sets of deaths is well-defined.
\cref{fig:approxwass} illustrates the exact computation of the 1-Wasserstein distance for connected components between four-node networks.

\subsection{Vector Representation of Persistence Barcodes}
\label{sec:vector_representation}

A collection of 1-dimensional persistence barcodes together with the Wasserstein distance is a metric space.
1-dimensional persistence barcodes can be embedded into a vector space that preserves the Wasserstein metric on the original space of persistence barcodes as follows.
Let $G_1,G_2,...,G_N$ be $N$ observed networks possibly with different node sizes. Let $F^{-1}_{G_i,B}$ be a pseudoinverse of network $G_i$.
The vector representation of a persistence barcode for connected components in network $G_i$ is defined as a vector of the pseudoinverse sampled at $1/m,2/m,...,m/m$:
\begin{equation}
    \bm{v}_{B,i} := \big( F^{-1}_{G_i,B}(1/m),F^{-1}_{G_i,B}(2/m),...,F^{-1}_{G_i,B}(m/m) \big)^\top .
\end{equation}
A collection of these vectors $M_B=\{\bm{v}_{B,i}\}_{i=1}^N$ with the $p$-norm $||\cdot||_p$ induces the $p$-norm metric $d_{p,B}$ given by
\begin{align}
    d_{p,B}(\bm{v}_{B,i},\bm{v}_{B,j}) &= ||\bm{v}_{B,i}-\bm{v}_{B,j}||_p \\
\label{eq:vecbwass}
                &= m \widehat W_{p,B}.
\end{align}
Thus, for $p=1$ the proposed vector space describes Manhattan distance, $p=2$ Euclidean distance, and $p \rightarrow \infty$ the maximum metric, which in turn correspond to the earth mover's distance ($W_1$) \citep{rubner2000earth}, $2$-Wasserstein distance ($W_2$), and the bottleneck distance ($W_\infty$) \citep{kerber2017geometry}, respectively, in the original space of persistence barcodes.
Similarly, we can define a vector space of persistence barcodes for cycles $M_D=\{\bm{v}_{D,i}\}_{i=1}^N$ with the $p$-norm metric $d_{p,D}$. The normed vector space $(M_B,d_{p,B})$ describes topological space of connected components in networks, while $(M_D,d_{p,D})$ describes topological space of cycles in networks.

The topology of a network viewed as a 1-skeleton is completely characterized by connected components and cycles.
Thus, we can fully describe the network topology using both $M_B$ and $M_D$ as follows.
Let $M_B \times M_D=\{(\bm{v}_{B,i},\bm{v}_{D,i})\,|\,\bm{v}_{B,i} \in M_B, \bm{v}_{D,i} \in M_D \}$ be the Cartesian product between $M_B$ and $M_D$ so the vectors in $M_B \times M_D$ are the concatenations of $\bm{v}_{B,i}$ and $\bm{v}_{D,i}$. For this product space to represent meaningful topology of network $G_i$, the vectors $\bm{v}_{B,i}$ and $\bm{v}_{D,i}$ must be a network decomposition, as discussed in \cref{sec:decomposition}. Thus $\bm{v}_{B,i}$ and $\bm{v}_{D,i}$ are constructed by sampling their psudoinverses with $m=\mathcal{V}-1$ and $n=1 + \frac{\mathcal{V} (\mathcal{V} - 3)}{2}$, respectively, where $\mathcal{V}$ is a free parameter indicating a reference network size. 
The metrics $d_{p,B}$ and $d_{p,D}$ can be put together to form a $p$-product metric $d_{p,\times}$ on $M_B \times M_D$ as \citep{deza2009encyclopedia}
\begin{align}
    d_{p,\times}\big((\bm{v}_{B,i},\bm{v}_{D,i}), (\bm{v}_{B,j},\bm{v}_{D,j})\big) &= \big([d_{p,B}(\bm{v}_{B,i},\bm{v}_{B,j})]^p + [d_{p,D}(\bm{v}_{D,i},\bm{v}_{D,j})]^p\big)^{1/p} \\
    \label{eq:sumofbarcodes}
    &= \big( [m \widehat W_{p,B}]^p + [n \widehat W_{p,D}]^p \big)^{1/p},
\end{align}
where $(\bm{v}_{B,i},\bm{v}_{D,i}), (\bm{v}_{B,j},\bm{v}_{D,j}) \in M_B \times M_D$, $m=\mathcal{V}-1$ and $n=1 + \frac{\mathcal{V} (\mathcal{V} - 3)}{2}$. Thus, $d_{p,\times}$ is a weighted combination of $p$-Wasserstein distances, and is simply the $p$-norm metric between vectors constructed by concatenating $\bm{v}_{B,i}$ and $\bm{v}_{D,i}$.
The normed vector space $(M_B \times M_D,d_{p,\times})$ is termed \emph{topological vector space} (TopVS).
Note the form of $d_{p,\times}$ given in (\ref{eq:sumofbarcodes}) results in an unnormalized mass after multiplying $m$ and $n$ by their reciprocals given in (\ref{eq:1dwasscc}) and (\ref{eq:1dwasscycle}). This unnormalized variant of Wasserstein distance is widely used in both theory \citep{cohen2010lipschitz,skraba2020wasserstein} and application \citep{carriere2017sliced,hu2019topology,songdechakraiwut2021topological} of persistent homology.
A direct consequence of the equality given in (\ref{eq:sumofbarcodes}) is that the mean of persistence barcodes under the approximated Wasserstein distance \citep{rabin2011wasserstein} is equivalent to the sample mean vector in TopVS.
In addition, the proposed vector representation is highly interpretable because persistence barcodes can be easily reconstructed from vectors by separating sorted births and deaths.

For a special case in which networks $G_1,G_2,...,G_N$ have the same number of nodes, the vectors $\bm{v}_{B,i}$ and $\bm{v}_{D,i}$ are simply the original birth set $B(G_i)$ and death set $D(G_i)$, respectively, and the $p$-norm metric $d_{p,\times}$ is expressed in terms of exact Wasserstein distances as
\begin{equation}
    d_{p,\times} = ( [m W_{p,B}]^p + [n W_{p,D}]^p )^{1/p}.
\end{equation}

\section{Validation using Simulated Networks}
\label{sec:sim}

Simulated networks of different topological structure are used to compare the classification performance of the proposed TopVS relative to that of several other methods.
While nearly any classifier may be used with TopVS, here we illustrate results using the $C$-support vector machine (SVM) \citep{chang2011libsvm} with the linear kernel. When the TopVS method is applied to different-size networks, we compute birth and death sets of the largest network as discussed in \cref{sec:decomposition}, and upsample birth and death sets of smaller networks to match that of the largest network in size. If the networks considered have the same size, we simply vectorize their birth and death sets.

The performance of TopVS is compared to five other methods published in the literature. Three of these methods are based on 2-dimensional persistence barcodes: the \emph{Persistence Image} (PI) vectorization \citep{adams2017persistence}, the \emph{Sliced Wasserstein} kernel (SWK) \citep{carriere2017sliced} and the \emph{Persistence Weighted Gaussian} kernel (PWGK) \citep{kusano2016persistence}. The other two benchmark methods are based on graph kernels: the \emph{Propagation} kernel (Prop) \citep{neumann2016propagation} and the \emph{GraphHopper} kernel (GHK) \citep{feragen2013scalable}. The PI method embeds persistence barcodes into a vector space in which classification is performed using linear SVMs. The SWK, PWGK, Prop and GHK methods are combined with SVMs using the kernel trick \citep{hofmann2008kernel} to perform classification.

The three persistence barcode methods require computation of 2-dimensional persistence barcodes from networks. We compute a 2-dimensional persistence barcode using the approach of \citet{otter2017roadmap} in which edge weights are inverted via the function $f(w)=1/(1+w)$. Then a point cloud is obtained from the shortest path distance between nodes.  Finally, the Ripser implementation \citep{ctralie2018ripser} of the Rips filtration \citep{ghrist2008barcodes} generates the persistence barcode from the point cloud.

The two graph kernel methods require node continuous attributes. We follow the experimental protocol of \citet{borgwardt2020graph} in which a node attribute is set to the sum of edge weights incident to the node. 

Implementation details of the baseline methods are provided in \cref{supp:implementation}.

\paragraph{Evaluation and tuning protocol}
Nested cross validation (CV) is used for selection of optimal hyperparameters and assessment of generalization capacity of the candidate algorithms for classifying networks.
Nested CV comprises an outer loop of stratified 2-fold CV and an inner loop of stratified 5-fold CV. The folds in stratified CV are selected by preserving the percentage of network samples for each group label. The inner loop is used to tune hyperparameters via grid search \citep{bergstra2012random} to determine the set of optimal hyperparameters that achieves the highest accuracy. The outer loop provides an unbiased performance evaluation for the model trained using the optimal hyperparameters from the inner loop. Thus, the nested CV procedure finds the average of accuracy scores over 2 folds in the outer loop using a model trained by the optimal hyperparameters obtained from the inner loop. Additional details on hyperparameter values and tuning are provided in \cref{supp:implementation}.

\begin{figure}
\centering
\centerline{\includegraphics[width=.55\linewidth]{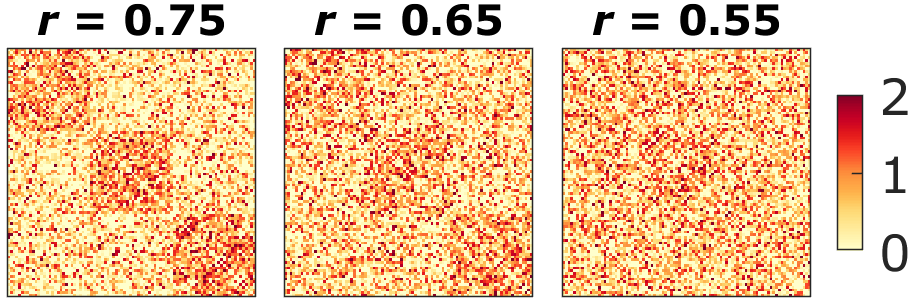}}
\caption{Example networks with $|V|=90$ nodes distributed evenly among $m=3$ modules according to within-module-connection probabilities $r = 0.75,0.65$ and $0.55$.}
\label{fig:modular_net}
\end{figure}

\paragraph{Simulated modular network structure}
Random modular networks $\mathcal{X}_i$ are simulated with $|V|$ nodes and $m$ modules such that the nodes are evenly distributed among modules. \cref{fig:modular_net} displays modular networks with $|V|=90$ nodes and $m=3$ modules such that $|V|/m=30$ nodes are in each module.
Edges connecting two nodes within the same module are assigned a random weight following a normal distribution $\mathcal{N}(1,0.5^2)$ with probability $r$ or otherwise Gaussian noise $\mathcal{N}(0,0.5^2)$ with probability $1-r$.
On the other hand, edges connecting nodes in different modules have probability $1-r$ of being $\mathcal{N}(1,0.5^2)$ and probability $r$ of being $\mathcal{N}(0,0.5^2)$. The modular structure becomes more pronounced as the within-module connection probability $r$ increases. Any negative edge weights are set to zero. This procedure yields random networks $\mathcal{X}_i$ that exhibit topological connectedness.

\paragraph{Simulated dataset}
Two groups of modular networks $L_1 = \{\mathcal{X}_i\}_{i=1}^{30}$ and $L_2 = \{\mathcal{X}_i\}_{i=31}^{60}$ corresponding to $m = 3$ and $5$ modules, respectively, are generated.
This results in 60 networks in the dataset, each of which has a group label $L_1$ or $L_2$.
Two different settings of network sizes are considered: 1) all 60 networks with $|V| = 90$ and 2) an equal number (ten) of networks with $|V| = 60, 90$ and 120 in each group. 
Three different settings of within-module connection probabilities are considered for each case: $r = 0.75,0.65$ and $0.55$ to vary the strength of the modular structure, as illustrated in \cref{fig:modular_net}.

\paragraph{Classification performance evaluation}
Binary classification is performed on the generated dataset using the candidate algorithms.
Nested CV is used to evaluate classification performance, resulting in an observed accuracy statistic $s$. 
Since the distribution of the accuracy $s$ is unknown, a permutation test is used to determine the empirical distribution under the null hypothesis that sample networks and their group labels are independent \citep{ojala2010permutation}. The empirical distribution is calculated by repeatedly shuffling the group labels, thereby removing any dependency between the sample networks and the labels, and then re-computing the corresponding nested CV accuracy score for one thousand random permutations. 
By comparing the observed accuracy to this empirical distribution, we can determine the statistical significance of the observed accuracy. 
The $p$-value is calculated as the fraction of permutations that give nested CV accuracy values higher than the observed accuracy $s$.
The average $p$-value and average observed accuracy across ten independently generated datasets are reported.

\begin{figure}[t]
\centering
\centerline{\includegraphics[width=1.15\linewidth]{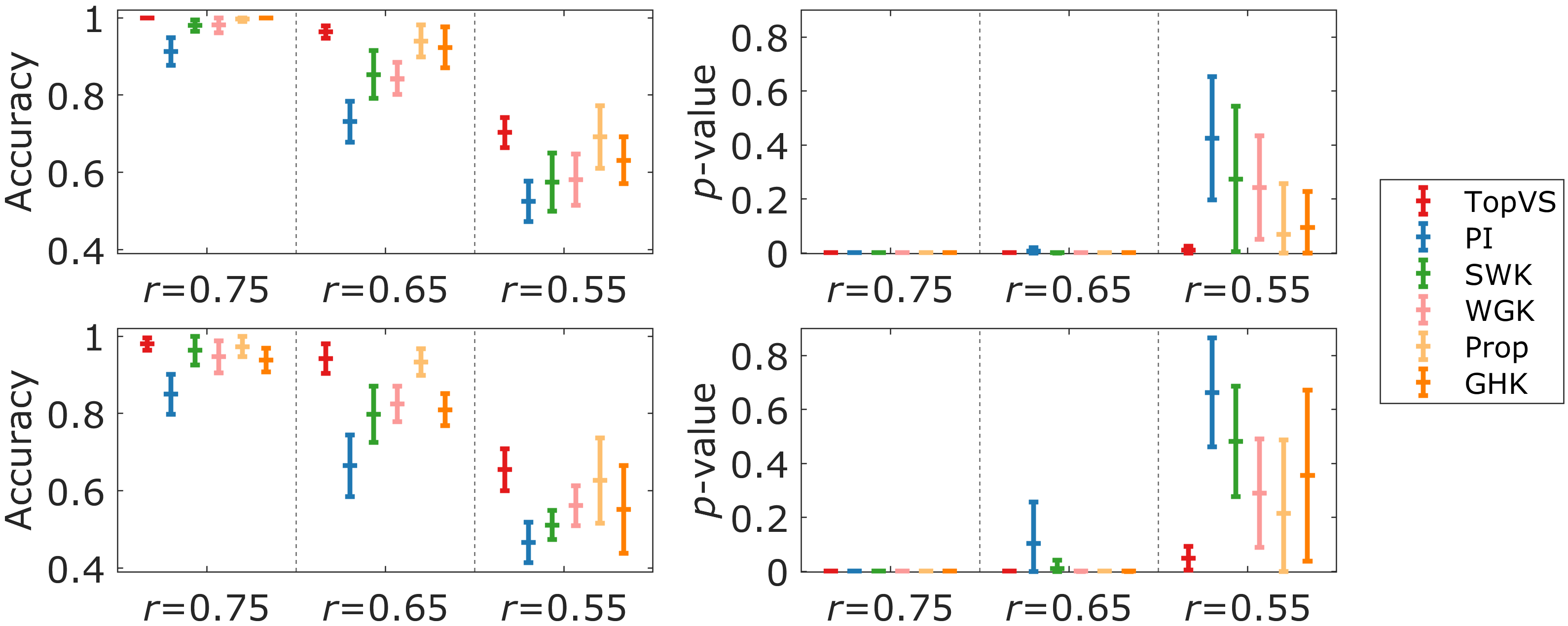}}
\caption{Classification performance comparison for simulated networks with $m = 3, 5$ modules and either $|V|=90$ nodes (Top) or $|V| = 60, 90, 120$ nodes (bottom) with respect to average accuracy (left) and average $p$-values (right). Results for within-module connection probabilities $r = 0.55,0.65$ and $0.75$ are shown. Data points (middle horizontal lines) indicate the average results over ten independently generated datasets, while vertical error bars indicate standard deviations.}
\label{fig:validation}
\end{figure}

\paragraph{Results}
\cref{fig:validation} indicates that all methods achieve relatively high accuracy on networks with pronounced modular structure ($r=0.75$), and their accuracy decreases as the modularity strength diminishes, i.e., decreasing $r$.
Our TopVS performs relatively well discriminating the more subtle modularity corresponding to $r=0.65$ and 0.55.
Since the dataset is purposefully generated to exhibit dependency between sample networks and their group labels, a low $p$-value provides statistical evidence that a trained classifier is able to leverage the dependency to differentiate network topology \citep{ojala2010permutation}. The proposed method has average $p$-values lower than 0.05 for all experimental settings, indicating that its improved accuracy over the baseline methods is significant. The Prop method has the closest accuracy to TopVS when $r = 0.55$, but has a higher $p$-value, indicating the accuracy is a less reliable indicator of performance.

\section{Application to Functional Brain Networks}
\label{sec:brain}

\begin{figure}
\centering
\centerline{\includegraphics[width=1.15\linewidth]{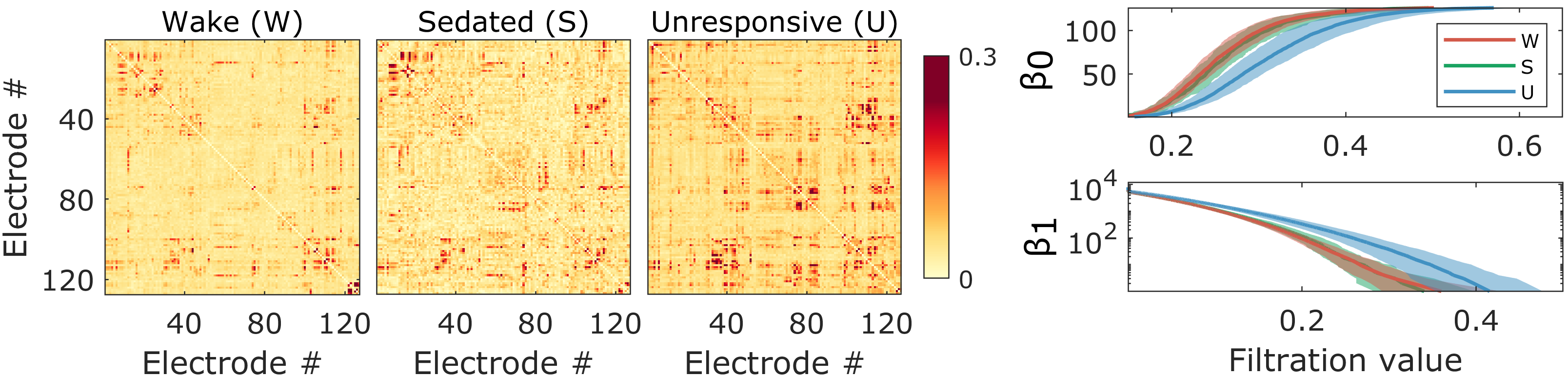}}
\caption{Data visualization of an example subject (ID L405). Left: Sample mean networks of wake, sedated, and unresponsive brains. Right: Network topology is completely characterized by connected component count $\beta_0$ and cycle count $\beta_1$. A thick line representing the mean of persistence barcodes and shaded areas around the mean representing standard deviation.}
\label{fig:brain_subject}
\end{figure}

\paragraph{Dataset}
We evaluate our method using an extended brain network dataset from the anesthesia study reported by \citet{banks2020cortical} (see \cref{supp:brain} for details). 
The measured brain networks are based on alpha band (8-12 Hz) weighted phase lag index \citep{vinck2011improved} applied to 10-second segments of resting state intracranial electroencephalography recordings. These recordings were made from eleven neurosurgical patients during administration of increasing doses of the general anesthetic propofol just prior to surgery.
Each segment is labeled as one of the three arousal states: pre-drug \emph{wake} (W),  \emph{sedated} but responsive to command (S), or \emph{unresponsive} (U). The number of networks (10-second segments) per subject varies from 71 to 119. The network size varies from 89 to 199 nodes across subjects. \cref{fig:brain_subject} illustrates sample mean networks and 1-dimensional persistence barcodes describing topology for a representative subject.

\paragraph{Classification performance evaluation}
All candidate methods used in the simulation study are evaluated on the brain network dataset.
We are interested in whether 1) the candidate methods can differentiate arousal states within individual subjects, and 2) generalize their learned knowledge to unknown subjects afterwards. As a result, we consider two different nested CV tasks as follows.

For the first task, we apply a nested CV comprising an outer loop of stratified 2-fold CV and an inner loop of stratified 3-fold CV, for each subject. Since we may get a different split of data folds each time, we perform the nested CV for 100 trials and report an average accuracy score and standard deviation for each subject. We also average these individual accuracy scores across subjects ($11 \times 100$ scores) to obtain an overall accuracy. 

For the second task, we use a different nested CV comprising both outer and inner loops with a leave-one-subject-out scheme. That is, a classifier is trained using all but one test subject. The inner loop is used to determine optimal hyperparameters, while the outer loop is used to assess generalization capacity of the candidate methods to unknown subjects in the population.

\begin{figure}
\centering
\centerline{\includegraphics[width=1.15\linewidth]{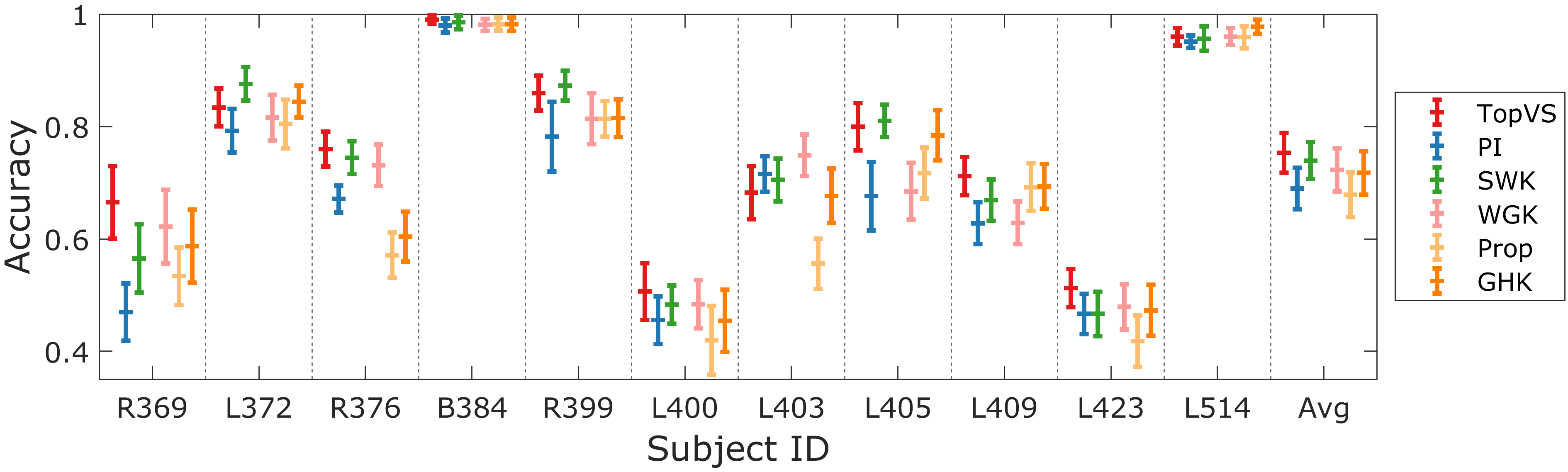}}
\caption{Accuracy classifying brain networks within individual subjects. The last column displays the average accuracy obtained across all subjects. The center markers and bars depict the means and standard deviations obtained over 100 different trials.}
\label{fig:brain_individual}
\end{figure}

\paragraph{Results}
\cref{fig:brain_individual} compares classification accuracy for individual subjects. There is variability in performance across subjects and across methods. In most subjects all methods perform relatively well. Our TopVS method is consistently among the best performing classifiers, resulting in the higher overall performance. On the other hand, the PI and Prop methods perform poorest in most subjects. The consistently poorer performance of PI and Prop is evident in the lower overall performance. Although the Prop method is among the best performing methods for the simulated networks, here it suffers when presented with functional brain networks.

\begin{figure}
\centering
\centerline{\includegraphics[width=1.15\linewidth]{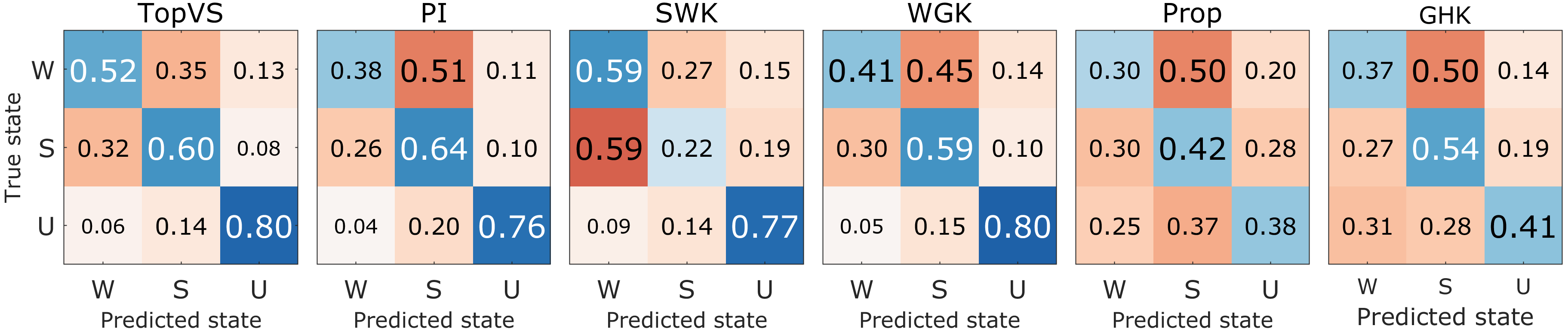}}
\caption{Confusion matrices illustrating method performance for classifying across subjects.
The numbers represent the fraction of brain networks in the test subjects being predicted as one of the three possible states: wake (W), sedated (S), and unresponsive (U). 
The confusion matrices are normalized with the entries in each row summing to 1.
}
\label{fig:conf_mtx}
\end{figure}

\cref{tab:brain_population} shows that TopVS is also among the best methods for classifying across subjects, while the performance of both graph kernel methods suffers.
\cref{fig:conf_mtx} displays a summary of the across-subject prediction results using confusion matrices. Except for two graph kernels, the other methods are generally effective for separating unresponsive (U) from the other two states. However, the majority of classification errors are associated with the differentiation between wake (W) and sedated (S) states. This misclassification is consistent with prior biological expectations since the sedated brain, in which subjects have been administered propofol but are still conscious, is expected to have a great deal of similarity with the wake brain \citep{banks2020cortical}. TopVS appears to show clear advantages over other baseline methods for differentiating wake and sedated states. This suggests that the proposed vector representation is an effective choice for representing subtle topological structure in networks.

\begin{table}
\caption{Mean accuracy and standard deviation for classifying brain networks of test subjects not used for training.}
\label{tab:brain_population}
\begin{center}
\begin{sc}
\begin{tabular}{cccccc}
\toprule
TopVS & PI & SWK & WGK & Prop & GHK \\
\midrule
$0.65 \pm 0.21$ & $0.58 \pm 0.22$ & $0.57 \pm 0.20$ & $0.60 \pm 0.21$ & $0.36 \pm 0.12$ & $0.43 \pm 0.14$ \\
\bottomrule
\end{tabular}
\end{sc}
\end{center}
\end{table}

\section{Conclusion}
\label{sec:conclusion}

TopVS is a vector representation for 1-dimensional persistence barcodes of connected components and cycles in complex networks.
The computational practicality of TopVS extends its applicability to the large-scale analyses of complex networks that cannot be analyzed using prior methods based on 2-dimensional persistence barcodes.
In addition, TopVS is a discriminative representation of complex networks with sound theoretical grounding based on the Wasserstein distance of the underlying barcode space. Our simulated and human-brain network studies demonstrate that TopVS performs well across a wide variety of classification scenarios.
Lastly, TopVS is highly interpretable and allows the computation of summary statistics for statistical inference.
The practical computation, discriminative power and high interpretability of TopVS will have a high impact on the analyses of large and complex network representations.

\subsubsection*{Acknowledgments}
This work supported in part by the National Institute of General Medical Sciences under award R01 GM109086 and the Lynn H. Matthias Professorship from the University of Wisconsin.

\bibliography{reference.bib}
\bibliographystyle{abbrvnat}

% APPENDIX

\newpage
\appendix
\onecolumn

\section{Implementation Details of Candidate Methods}
\label{supp:implementation}

Implementation of \emph{Persistence Image} (PI) vectorization \citep{adams2017persistence} is performed using Persim \citep{scikittda2019}. \emph{Sliced Wasserstein} kernel (SWK) \citep{carriere2017sliced} and \emph{Persistence weighted Gaussian} kernel (PWGK) \citep{kusano2016persistence} are implemented using the Gudhi library \citep{gudhiproj2022gudhi}. \emph{Propagation} kernel (Prop) \citep{neumann2016propagation} and \emph{GraphHopper} kernel (GHK) \citep{feragen2013scalable} are implemented via GraKel library \citep{siglidis2020grakel}.

For the PI method, we follow a parameter setting used by \citet{adams2017persistence} to convert two persistence barcodes for connected components and cycles into two 2-dimensional pixel images of $20 \times 20$ resolution using a Gaussian function with variance $0.01$. The two images are vectorized and concatenated into a single feature vector per network. Then linear SVMs are used to classify these vectors.

SWK is based on the sliced Wasserstein approximation \citep{rabin2011wasserstein} over $10$ directions. PWGK is based on the RBF kernel and an $\arctan$ weight function recommended by \citet{kusano2016persistence}.
Both SWK and PWGK use combined persistence barcodes, each comprising 2-dimensional points of both connected components and cycles, to compute the Gram matrices.

Grid search \citep{bergstra2012random} across different hyperparameter values is used to train all the candidate methods. SVMs have a regularization parameter $\mathcal{C}=\{0.01,1,100\}$. Thus, a grid search trains TopVS and PI methods with each $C\in\mathcal{C}$. The SWK and WGK methods have a bandwidth parameter $\Sigma=\{0.1,1,10\}$, and thus grid search trains both methods with each pair $(C,\sigma) \in \mathcal{C} \times \Sigma$. The Prop method has a maximum number of propagation iterations $T_{max}=\{1,5,10\}$, and thus is trained with each pair $(C,t_{max}) \in \mathcal{C} \times T_{max}$. GHK method uses the RBF kernel with a parameter $\Gamma=\{0.1,1,10\}$ between node attributes, and thus is trained with each pair $(C,\gamma) \in \mathcal{C} \times \Gamma$.

\section{Brain Network Dataset}
\label{supp:brain}
Brain network data were obtained from eleven neurosurgical patients between 19 and 59 years old as described in \cref{supptab:dataset}. The patients were undergoing chronic invasive intracranial electroencephalography (iEEG) monitoring as part of their treatment for medically refractory epilepsy. The Code of Ethics of the World Medical Association (Declaration of Helsinki) for experiments involving humans was followed for all the experiments. The  University of Iowa Institutional Review Board and the National Institutes of Health approved all research protocols, and written informed consent was obtained from all subjects. Acquisition of clinically required data was not impeded by the research and subjects were free to rescind their consent whenever they wished without interfering with their clinical evaluation. Subdural and depth electrodes (Ad-Tech Medical, Oak Creek, WI) used to obtain all research data were located by the team of epileptologists and neurosurgeons based solely on needs for clinical evaluation of  the patients. Data collected in the operating room prior to electrode removal, before and during induction of general anesthesia with propofol were used to create the brain network dataset. Full description of the method for obtaining the brain network dataset and experimental procedure is provided in \citep{banks2020cortical}.

\begin{table}[h]
\caption{Brain network dataset.}
\label{supptab:dataset}
\centering
\vskip 0.15in
\begin{sc}
\begin{tabular}{lccc}
    \toprule
    Subject & Age & Gender & Network size  \\
    \midrule
    R369 & 30 & M & 199 \\
    L372 & 34 & M & 174 \\
    R376 & 48 & F & 189 \\
    B384 & 38 & M & 89 \\
    R399 & 22 & F & 175 \\
    L400 & 59 & F & 126 \\
    L403 & 56 & F & 194 \\
    L405 & 19 & M & 127 \\
    L409 & 31 & F & 160 \\
    L423 & 51 & M & 152 \\
    L514 & 46 & M & 118 \\
    \bottomrule
\end{tabular}
\end{sc}
\end{table}

\end{document}